%% file: ms.tex
\documentclass[sigconf,authorversion,nonacm]{acmart}

\AtBeginDocument{%
  \providecommand\BibTeX{{%
    \normalfont B\kern-0.5em{\scshape i\kern-0.25em b}\kern-0.8em\TeX}}}

\usepackage{graphicx}
\graphicspath{ {./images/} }

\usepackage{amsmath}

\usepackage[ruled,vlined]{algorithm2e}
\newcommand{\nosemic}{\renewcommand{\@endalgocfline}{\relax}}
\usepackage{tabularx}

\newcolumntype{Y}{>{\centering\arraybackslash}X}

\DeclareMathAlphabet{\mathds}{U}{BOONDOX-ds}{m}{n}
\DeclareMathOperator*{\argmax}{arg\,max}

\newcommand{\R}{\mathbb{R}}
\newcommand{\E}{\mathbb{E}}

\acmConference[UrbComp 2021]{The 10th International Workshop on Urban Computing}{November 1, 2021}{Beijing, China}

\begin{document}

\title{Back to Basics: Deep~Reinforcement~Learning~in~Traffic~Signal~Control}

\author{Sierk Kanis}

\email{sierkkanis@hotmail.com}


\affiliation{
  \institution{University of Amsterdam}
  \city{Amsterdam}
  \country{The Netherlands}}

\author{Laurens Samson}
\email{l.samson@amsterdam.nl}

\author{Daan Bloembergen}
\email{d.bloembergen@amsterdam.nl}
\affiliation{%
  \institution{CTO, City of Amsterdam}
  \city{Amsterdam}
  \country{The Netherlands}
}

\author{Tim Bakker}
\email{t.b.bakker@uva.nl}
\affiliation{%
 \institution{University of Amsterdam}
 \city{Amsterdam}
 \country{The Netherlands}}


\begin{abstract}
In this paper we revisit some of the fundamental premises for a reinforcement learning (RL) approach to self-learning traffic lights. We propose RLight, a combination of choices that offers robust performance and good generalization to unseen traffic flows. In particular, our main contributions are threefold: our lightweight and cluster-aware state representation 
leads to improved performance; 
we reformulate the Markov Decision Process (MDP) such that it skips redundant timesteps of yellow light, speeding up learning by 30\%; and we investigate the action space and provide insight into the difference in performance between acyclic and cyclic phase transitions. 
Additionally, we provide insights into the generalisation of the methods to unseen traffic.
Evaluations using the real-world Hangzhou traffic dataset show that RLight outperforms state-of-the-art rule-based and deep reinforcement learning algorithms, demonstrating the potential of RL-based methods to improve urban traffic flows.
\end{abstract}





\maketitle

\section{Introduction}
\input{chapters/introduction}

\section{Background}
\label{background}
\input{chapters/background}

\section{Related Work}
\label{literature_review}
\input{chapters/literature_review}

\section{RL{\small{ight}} Agent Design}
\label{agent_design}
\input{chapters/agent_design}

\section{Self-Organizing Traffic Lights 2.0}
\label{chapter:sotl}
\input{chapters/methods}

\section{Experiment Setup}
\label{results}
\input{chapters/results}

\section{Discussion}
\label{discussion}
\input{chapters/discussion}

\section{Conclusion}
\label{conclusion}
\input{chapters/conclusion}

\bibliographystyle{ACM-Reference-Format} \balance
\bibliography{Additions/Thesis.bib}

\end{document}

%% file: chapters/introduction.tex
Adaptive Traffic Signal Control (ATSC) aims to facilitate smooth and safe traffic flow, thereby reducing travel times and cutting CO$_2$ emissions~\cite{zhao2011computational}. While conventional methods rely on handcrafted rules based upon specific traffic scenarios, the recent increase in traffic data and enormous advances in optimization techniques suggest that more opportunities are at hand \cite{WEI2019survey}.

In recent years, combining deep learning with reinforcement
learning (RL) has led to strong performance in many complex environments, often achieving super-human performance \cite{mnih2015human,silver2018general,hernandez2019survey}. These methods have proven to be able to handle dynamic environments, starting with a blank slate and learning directly from feedback by trial and error without relying on simplistic data assumptions. It is natural to wonder whether deep RL could be just as beneficial for ATSC.

However, ATSC presents a challenge from an RL perspective. While in some RL applications the environment straightforwardly translates to a Markov Decision Process (MDP), ATSC has no principled source of raw data and rewards, while also lacking a fixed set of actions or action-rate. This means that success is highly reliant on the quality of the MDP representation.

\begin{figure*}[tb]
  \includegraphics[height=4.3cm]{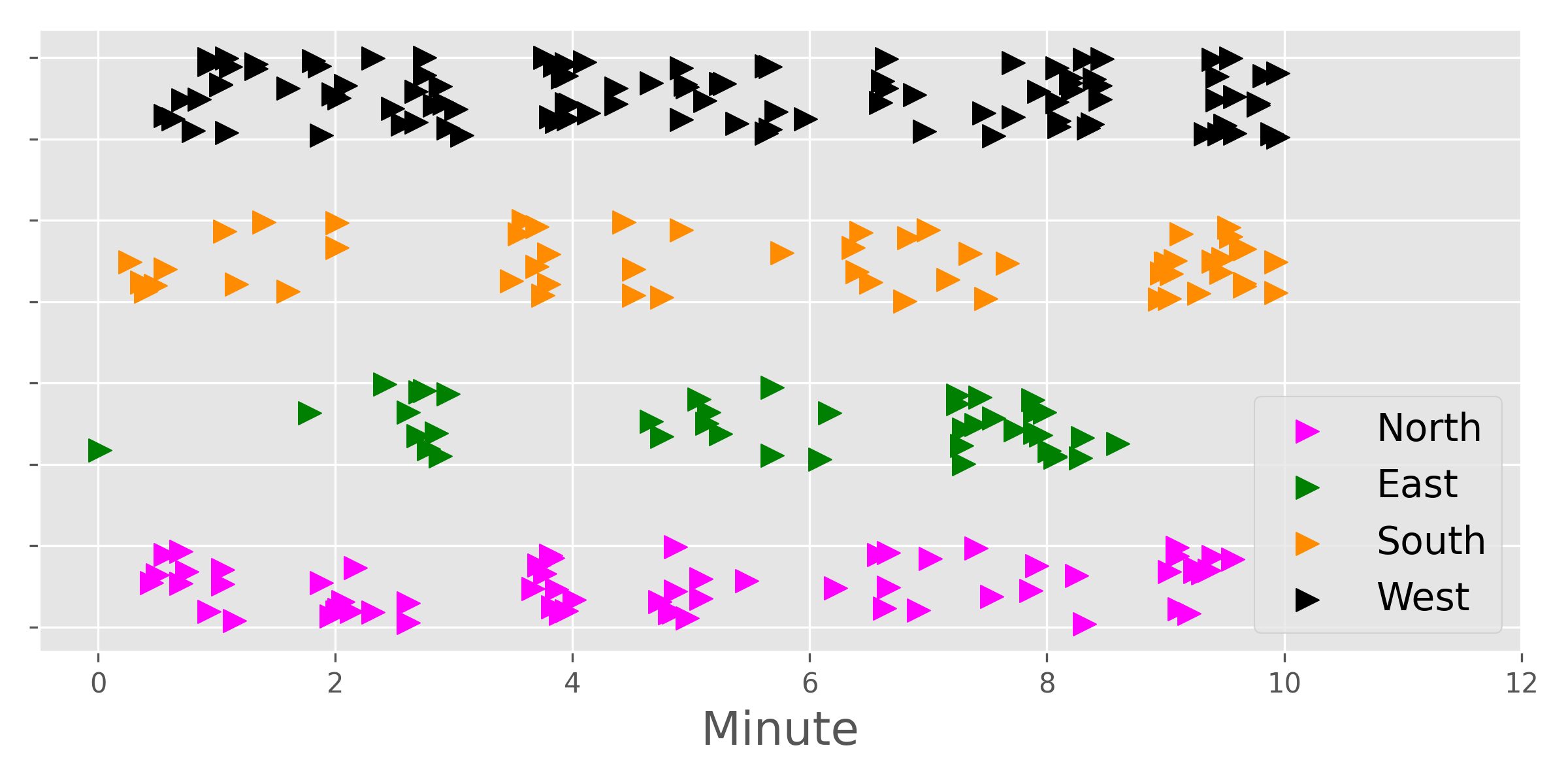}
  ~
  \includegraphics[height=4.3cm,trim={0.85cm 0.65cm 0cm 0cm},clip]{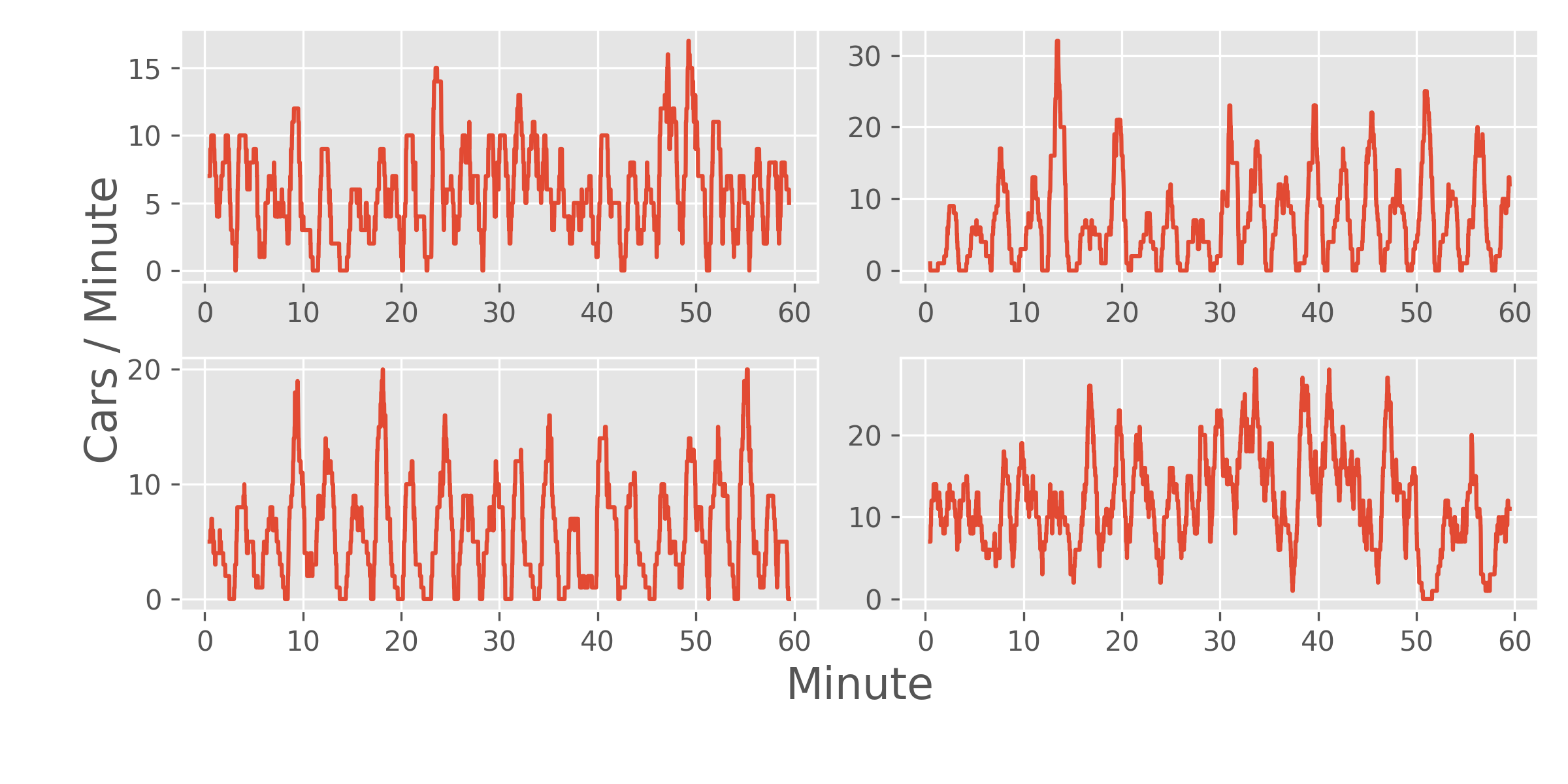}
\caption{The left panel shows data from the first ten minutes of inflow on intersection Hangzhou 2, with every triangle indicating a vehicle. The right panel shows the rate of vehicles per minute. Notice how traffic often appears in clusters.}
\label{fig:clusters}
\end{figure*}

Models that aim to equip the agent with all possible information are unnecessarily complex which consequently hurts performance~\cite{zheng2019diagnosing}. On the contrary, Light-Intellight (LIT) \cite{zheng2019diagnosing} aims to propose a minimal set of features based upon a uniform traffic distribution and shows good performance. However, traffic distributions in dense urban areas are often non-uniform and consist of clusters, see Figure~\ref{fig:clusters}. No prior work seems to specifically design their state representation to fit clustered traffic data.


We propose Reinforcement-Learning-Light (RLight)\footnote{\url{https://github.com/Amsterdam-Internships/Self-Learning-Traffic-Lights}}, a simple yet concise RL framework that exploits inductive biases relevant to the ATSC problem. We build upon the advantages of LIT using a simple shaping reward, yet expand their state representation to fit a more clustered data distribution. We represent approaching and waiting vehicles separately and add the mean speed and distance of the approaching cluster to the state as an approximation, thereby exploiting the specific structure of clusters to keep the state representation compact and digestible. By using only structured data, our method can exploit the current sensors in the ground and GPS data, therefore contributing to the transition of adopting a reinforcement learning approach in the real world.

Additionally, to our knowledge, no prior work has acknowledged the ambiguity in modelling yellow light. So far the ATSC problem has mainly been formulated as a standard MDP, whereas an underlying inductive bias could be built in to skip redundant timesteps during yellow light, thus cleaning up the reward signal by ignoring irrelevant state-action pairs. We show that this reduces learning time by approximately one third.

Furthermore, prior work does not agree on whether it is safe to have an acyclic action space or not, whereas we find that limiting the agent to cyclic control impacts the performance significantly. Just as a human traffic warden would do, we think it is more natural to be able to freely choose the optimal phase at every timestep. We investigate this difference to provide more ground for utilizing acyclic control.

We apply our approach to real data of five intersections in the city of Hangzhou using the CityFlow simulator \cite{zhang2019cityflow}. Since this dataset is fixed and thus, in RL terms, the environment is deterministic, we split up the data into train, validation and test sets to provide insights into the generalisation of our method. 
To test our approach against a rule-based method, we propose an extension of the
Self-Organizing Traffic Lights (SOTL) method \cite{Cools2008} that works in multi-phase settings as well as in the originally implemented two-phase setting. Empirically, we show that our proposed method RLight consistently outperforms state-of-the-art methods on all five intersections.

In short, our contributions are the following:
\begin{itemize}
    \item We propose RLight, a simple yet concise RL framework that exploits inductive biases relevant to the ATSC problem:
    \begin{itemize}
    \item Our lightweight and cluster-aware state representation outperforms state-of-the-art methods on the Hangzhou dataset.
    \item We reformulate the MDP such that it skips redundant timesteps of yellow light, speeding up learning.
    \item We investigate the action space and provide insight into the difference in performance between acyclic and cyclic phase transitions.
    \end{itemize}
    \item In addition, we extend the rule-based algorithm SOTL to work in an acyclic fashion for multi-phase intersections.
\end{itemize}

Our single agent is able to successfully control varying traffic densities, enabling it to generalize to unexpected real-world situations such as a football event, a nearby accident or, let's say, a world-wide pandemic.

%% file: chapters/background.tex
In this section we elaborate on the basics of traffic signal control, we give an overview of reinforcement- and deep reinforcement learning, and we go over the basics of the Self-Organising Traffic Lights (SOTL) algorithm.

\subsection{Traffic Signal Control}

We address the traffic signal control problem using a set of incoming lanes $\pmb{\nu}$ of cardinality $J$, a set of outgoing lanes and a set of phases $\pmb{\varphi}$ of cardinality $I$. Each phase is a combination of green and red lights and the set of phases consists of all possible combinations in which conflicting directions do not have green light simultaneously. In particular, this means every phase consists of two green lights, while the other lights are red. A fixed yellow period of five seconds is enforced after every switch of lights.
For simplicity, we assume that separating conflicting directions and forcing yellow light periods account for safety and we do not consider safety in more detail.



\subsection{Reinforcement Learning}

Reinforcement learning is a computational approach characterized by its learning from direct interaction with the environment without requiring supervision or a complete model of the environment~\cite{sutton2018reinforcement}.


\paragraph{Markov Decision Process}  We consider a Markov Decision Process (MDP), which is a formalization of sequential decision making where actions not only influence immediate rewards, but also subsequent states and through those future rewards. In an MDP, the agent and environment interact at each of a sequence of discrete timesteps, $t = 0, 1, ..., T$. At every timestep $t$, the agent receives a representation of the environment state $s_t$ and selects an action $a_t$, after which the environment transitions to the next state $s_{t+1}$
and the agent receives a reward $r_{t+1}$. The goal of reinforcement learning is to learn a policy, i.e. a mapping from perceived states to actions in those states $\pi: S \rightarrow \Delta_A$ that maximizes the expected return, which is defined as:

\[
Q_\pi(s,a) = \E_\pi \left[ \sum_{t \geq 0} \gamma^t r_t \middle| S_0 = s, A_0 = a\right]
\]

for each initial state-action pair $(s, a) \in S \times A$, where $\gamma \in [0, 1]$ is the time-discount factor that prioritises short-term rewards over long-term rewards \cite{sutton2018reinforcement}.

\paragraph{State} At each time step, the agent receives a quantitative representation of the environment, i.e. a state representation $S_t$. Ideally, the state representation fully describes the environment, such that the agent is given all information necessary to perform the right action. However, an overly complex state representation toughens distilling the useful information, while on the other hand, a simple state representation may result in different states appearing to be identical, making it impossible to learn the appropriate behaviour. In some reinforcement learning applications the environment straightforwardly translates to a state representation, e.g. pixels on a screen. In the traffic signal control setting, however, no principled source of raw data exists, which means success is highly reliant on the quality of the hand-crafted state representation.

\paragraph{Reward} On each timestep the environment sends the RL-agent a reward signal $R_t$ indicating the quality of its chosen action. This allows the agent to estimate a value function, i.e. the total amount of reward the agent can expect to accumulate over the future from each state. Subsequently, actions are taken based on this estimated value function, seeking to reach states with higher values. 

Again, there is no clear-cut reward signal in the traffic signal control setting, like points in Tetris or winning a game of chess. In the case of very sparse reward signals like average travel time, the addition of intermediate shaping rewards can be helpful to steer the agent towards the goal \cite{ng1999policy}. However, rewards must be provided in such a way that in maximizing them the agent will also achieve the goals one want to be accomplished.

\paragraph{Actions} Each timestep $t$ the agent selects an action $A_t$ from the action space $A = \{1, ..., K\}$, i.e. the set of possible actions. Ideally, the action space allows the agent maximum freedom in the environment, such that all options are available necessary to perform the right action. However, a too big action space imposes the agent with the task to choose the right action from a bigger set, possibly harming performance. In traffic signal control there is no fixed set of available options in the environment, e.g. the legal moves in the game of chess. This means that performance is reliant on the choice of the action space.

\subsection{Deep Reinforcement Learning}


Deep reinforcement learning (Deep RL) can be understood as consisting of three fundamental components: a function approximator, a learning algorithm and a mechanism for generating training data. We parameterize an approximate value function $Q(s,a;\theta_i)$ using a deep neural network architecture in which $\theta_i$ are the parameters of the network at iteration $i$. Samples of experiences generated by our traffic simulator $(s,a,r,s') \sim U(D)$ are drawn uniformly at random from the memory of saved transitions and are used for learning by updating the local neural network at iteration $i$ by the following loss function:

\[ L_i(\theta_i) = \E_{(s,a,r,s') \sim U(D)} \left[ r + \gamma \max_{a'} Q(s', a'; \theta_i^-) - Q(s, a; \theta_i) \right] \]

in which $\theta_i$ are the parameters of the local neural network at iteration i and $\theta_i^-$ are the parameters of the target network at iteration i \cite{mnih2015human}. The target network is added to stabilize learning and its parameters $\theta_i^-$ are being soft-updated with factor $\tau$ to slowly catch-up with the local network~\cite{lillicrap2015continuous}.




\subsection{Self-Organizing Traffic Lights}
\label{background:sotl}

Next to deep reinforcement learning approaches, we examine the rule-based Self-Organizing Traffic Lights (SOTL) algorithm, in order to compare our RLight agent to a rule-based approach. SOTL is a distributed adaptive traffic light system with no communication between intersections. The method is considered self-organizing, because the global performance is only dependent on the local rules of each intersection; each intersection is unaware of the state of other intersections~\cite{Cools2008}. Still, it is able to achieve global coordination of traffic because of the probabilistic formation of vehicle clusters~\cite{Gershenson2004}.

Each traffic phase $\pmb{\varphi}_i$ keeps a count $\kappa_i$, with $i$ from 1 to $I$, in which $I$ is the number of phases. At every time-step, $\kappa_i$ adds the number of vehicles on the currently red lanes of $\pmb{\varphi}_i$, independently of whether a vehicle is moving or waiting. When $\kappa_i$, representing the integral of vehicles on the lanes of $\pmb{\varphi}_i$ over time, reaches a threshold $\theta$, the current phase switches to phase $\pmb{\varphi}_i$ and $\kappa_i$ is reset to zero. To prevent fast switching of lights, additional constraints are tuned per intersection \cite{Gershenson2004}. The original implementation by \citeauthor{Gershenson2004} only supports two phases, in Section \ref{chapter:sotl} we extend the method to work in multi-phase settings too.

%% file: chapters/literature_review.tex
The ATSC problem has traditionally been approached using rule-based methods that use manually tuned parameters, as well as machine learning techniques \cite{zhao2011computational}.
Although approaches based on RL go back more than two decades \cite{wiering2000multi}, the advent of Deep RL has led to a new boom in research in this direction \cite{haydari2020deep}. Many RL approaches discuss various ways in which to define the underlying Markov Decision Process (MDP) in terms of its state, action and reward representation \cite{WEI2019survey}.

\paragraph{State} The introduction of deep learning has shown that training directly from raw inputs can lead to better representations than handcrafted features \cite{mnih2015human}, but the ATSC setting does not have a principled source of raw data. As an attempt to fully describe the traffic situation recent studies have used grids \cite{liang2019deep} or images \cite{Pol2016}, which led to state representations with thousands of dimensions. This abundance of information however does not necessarily lead to a gain in performance \cite{WEI2019survey}.

\paragraph{Reward} A principal goal of ATSC is to reduce the average travel time in the system. However, this metric is hard to estimate outside a simulator and additionally leads to highly delayed signals. Recent studies define the reward function as an ad-hoc weighted linear combination of several direct traffic measures such as queue length, waiting time, speed, and throughput \cite{Pol2016,mannion2016experimental,Wei2018}. Such multi-component reward exhibits two drawbacks. First, there is no guarantee that the desired goal is optimized. Second, small changes in the component weights could lead to drastically different results~\cite{zheng2019diagnosing}. Unfortunately, there is no principled approach to tune weights within a RL reward function \cite{WEI2019survey}.\\

The work closest to ours is Light-Intellight (LIT) \cite{zheng2019diagnosing}. Their approach uses the cumulative queue length as a proxy for average travel time. The authors show that queue length is proportional to travel time and therefore optimizing queue length amounts to optimizing travel time. With this reward, \citeauthor{zheng2019diagnosing} claim that only the number of vehicles and the traffic signal phase are needed to fully describe the system, under the assumption of uniform traffic inflow. Under this assumption their method worked well, but as shown in Figure \ref{fig:clusters}, urban areas with dense intersections cause traffic to be highly clustered. In this work, we adopt the reward function of \citeauthor{zheng2019diagnosing} but challenge their assumption of uniformity by expanding our state representation to adapt to more fragmented traffic distributions.

\paragraph{Action} The LIT approach, among others, additionally assumes that a predetermined phase order aligns best with drivers' expectations and avoids safety issues \cite{zheng2019diagnosing,liang2019deep}. Yet, there are urban traffic control systems that do utilize acyclic phases (e.g. Amsterdam).
Since limiting the agent to a fixed cycle restricts it to learn the optimal policy \cite{Genders2019}, we investigate the reduction of travel time when allowing acyclic phase transitions.

\paragraph{Coordination} In recent years, studies have scaled up their approach to large multi-agent systems, proposing multiple forms of coordination and communication between the agents \cite{mannion2016experimental,haydari2020deep,Chen2020}. \citet{Pol2016} have extended their single agent DQN solution by making use of transfer planning and the max-plus coordination algorithm while \citet{wei2019colight} have introduced a graph attentional network to facilitate cooperation. However, under certain circumstances, explicit coordination between intersections may not be a necessity \cite{girault2016exploratory}. Traffic lights themselves already structure traffic flow in clusters of vehicles. 
When intersections follow each other rapidly, clusters likely persist until the next intersection.
The large empty areas that appear between clusters could then be used by crossing clusters; if agents can learn to anticipate this approaching traffic, the system can naturally become self-organizing without the need for explicit coordination \cite{Cools2008}.
Empirically, \citet{zheng2019diagnosing} have shown that LIT outperforms the multi-agent approach of \citet{Pol2016} without adding explicit coordination, which motivates us to investigate the quality of the basic assumptions of single-agent control further.



%% file: chapters/agent_design.tex
Our RLight agent is based on the standard RL framework \cite{sutton2018reinforcement}. In the following, we describe how we formulate ATSC as a Markov Decision Process (MDP) and discuss the state and action representation, as well as the reward function. We consider the scenario in which the agent controls a single intersection with $J$ incoming lanes, and $I$ traffic light phases.

\subsection{Markov Decision Process}
\label{mdp_smdp}

To our knowledge, no prior work has investigated the difference in modelling the action rate in ATSC problems. Assuming a timestep rate of one simulation second per transition and a fixed period of yellow light enforced by the environment, we consider two options.

\paragraph{MDP} The agent selects an action at every simulation second and we accept that actions during yellow light are being ignored by the environment. This means that the agent would have to learn that actions in such states do not have any impact on the final reward, and therefore, every action is equally good. In this way, training effort is put into learning irrelevant state-action pairs, while it can also be prone to making the reward signal noisier. Consider the scenario when the agent has not fully learned to distinguish yellow and non-yellow states, in which case it appears to the agent as if it gets rewarded for its action, yet in a yellow state it would get this reward anyway.

\paragraph{SMDP} The agent only chooses actions when its decisions actually have an impact on the environment, making it inactive during yellow light. This can be seen as a Semi-Markov Decision Process (SMDP) \cite{sutton1999between}.
Assume that the yellow period has a fixed length of $\psi$ timesteps. Whenever the agent decides to switch to another phase (and thus the yellow light period starts) at timestep $\tau$, in state $s_\tau$ with action $a_{\tau}$, it receives the cumulative discounted rewards during the yellow period added to the discounted immediate reward after the yellow period:
\[
r_{\tau}(s_{\tau}, a_{\tau}) = \sum_{t = 1}^{\psi+1} \gamma^t r_{\tau + t}(s_{\tau + t})
\]
The advantage of this way of modelling is that irrelevant state-action pairs do not have to be learned, saving training time and simplifying optimization. Note that, since the queue length of each timestep is still implicitly perceived by the agent, the proportionality of queue length and average travel time as discussed in Section~\ref{reward_function} still holds.

\subsection{State representation}
\label{state_representation}
At each time step, the agent receives a quantitative representation of the environment, i.e. a state representation. Our goal is to choose a representation that is easily digestible, yet concise enough to contain the information necessary to select the appropriate actions.

As a starting point, we consider only the number of vehicles and the current phase as used in LIT \cite{zheng2019diagnosing}.
The current phase is necessary for the agent to know whether lights will switch by taking an action and the waiting and approaching vehicles are represented jointly based upon the idea that, if the inflow of traffic is constant, the agent can estimate these instead of perceiving them directly.

This leads to:
\begin{equation*}
    \pmb{s_t} = [\pmb{w}^\intercal_t + \pmb{a}^\intercal_t, \pmb{p}^\intercal_t],
\end{equation*}
in which $\pmb{w}$ + $\pmb{a} \in \R^J$ is a transposed vector 
representing the total number of vehicles (waiting, $\pmb{w}$, plus approaching, $\pmb{a}$) on each lane and $\pmb{p}$ is the phase represented as a one-hot vector of size $I$, where $I$ and $J$ are the number of phases and lanes, respectively. While LIT uses phase-gate \cite{wei2018intellilight} to condition the action-value function on the current state, we simply encode the phase as a one-hot vector and append it to the state.

However, not distinguishing approaching and waiting vehicles may result in different states appearing to be identical, making it impossible to learn the appropriate behaviour. In particular, this state representation cannot distinguish patterns in the traffic flow, and thus also cannot be used for implicit coordination between traffic lights.

We extend this approach by explicitly separating the waiting and approaching vehicles and thereby fitting the more fragmented distribution of urban traffic. 
Additionally, we concatenate the average speed and distance of the approaching traffic to improve traffic anticipation, i.e. switch lights earlier if a cluster moves faster or closer.

This results in the following state representation:
\begin{equation*}
    \pmb{s}_t = [\pmb{w}^\intercal_t, \pmb{a}^\intercal_t, \pmb{d}^\intercal_t, \pmb{s}_t^\intercal, \pmb{p}^\intercal_t]
\end{equation*}
in which $\pmb{w}$ represents the number of waiting vehicles, $\pmb{a}$ the number of approaching vehicles, $\pmb{d}$ the average distance of approaching vehicles, $\pmb{s}$ the average speed of approaching vehicles, and $\pmb{p}$ the phase represented as a one-hot vector as before. All values are normalized by dividing by their maximum observed value. We use the average speed and distance of the cluster as an approximation to keep the state representation compact, under the assumption that vehicles behave like steady convoys. In an eight-approach, twelve-phase setting $(J=8, I=12)$ this results in a state-space of dimension $4 \times 8 + 12 = 44$. Note that even under uniform traffic scenarios our expansion will be minimally as descriptive as LIT's approach.

\subsection{Action Space}

Each timestep $t$ the agent selects an action $a_t$ from the available set of actions $A = \{1, ..., K\}$. 
Our goal is to allow the agent maximum freedom, such that all necessary possibilities are available to select the right action. We investigate the following two options.

\paragraph{Cyclic} We use a predetermined phase sequence in which the agent can either keep the current phase or switch to the next phase. Specifically, the neural network outputs a value to switch or to keep.

\paragraph{Acyclic} We use an acylic phase sequence in which the agent can freely choose which phase to switch to next, allowing more flexible control. In this case, the network outputs as many values as there are phases.

\subsection{Reward Function}
\label{reward_function}

At every timestep the agent receives a numerical reward $r_{t}$ defined by the reward function. Our goal is to define a reward function to minimize average travel time which for simplicity we assume to be the main objective.

Since average travel time can only be computed at the end of a simulation trajectory, using this as the reward signal leads to extremely sparse and delayed rewards. In addition, the average travel time is generally not readily available outside a simulator. Following \citet{zheng2019diagnosing} we use the total queue length at the intersection as a shaping reward:
\[
r_t(s_t) = -\sum_{j=1}^{J} w_j(s_t)
\]
in which $w_j$ is the queue length at the lane $j$ and J is the number of lanes. This shaping reward is proven to be approximately proportional to the average travel time, when neglecting speed changes for simplicity \cite{zheng2019diagnosing}.

%% file: chapters/methods.tex


In this section, we explain how we generalize the rule-based Self-Organizing Traffic Lights (SOTL) algorithm to multi-phase settings.

The original method was only implemented for two-phase settings, which means it was implemented to switch to the next phase instead of explicitly switching to the phase with the most waiting time. In a two-phase setting, this amounts to the same policy, because in a two-phase setting cyclic and acyclic control is ambiguous. In a multi-phase setting, however, a phase should not only switch but also decide which phase to switch to. Also, in a multi-phase setting, multiple phases can get above the threshold simultaneously. Therefore we augment the algorithm by letting it choose the phase corresponding to the maximum $\kappa$ instead of the next phase in the sequence, such that the phase with the most cumulative waiting time is chosen. 

The original implementation brings along another issue when adapting it to a multi-phase intersection: Resetting $\kappa$ to zero is not anymore sufficiently informing the system that vehicles have passed through green, as a green light is part of multiple phases. Consider $\kappa_i$ and $\kappa_j$ $(i \neq j)$, which share one lane. If phase $\pmb{\varphi}_i$ is set to green, vehicles on the shared lane will pass through green, yet only $\kappa_i$ is reset, while $\kappa_j$ does not get updated. Consequently, the passed vehicles are still present in the integral of $\kappa_j$, which results in phase $\pmb{\varphi}_j$ being chosen next, although no vehicles are waiting anymore. Since SOTL uses counters to represent the integrals, there is no way to remove only the waiting time of one lane. This means that to keep functioning solely with counters (such that only one induction loop per lane is necessary) an additional parameter is called for.


\begin{algorithm}[tb]
\caption{SOTL generalized to multi-phase settings}
\label{sotl}
\DontPrintSemicolon
    initialize $\kappa$ and $\rho$\ to 0 \;
    \For{$t \gets 1$ \KwTo $T$}
    {
        \For{$j \gets 1$ \KwTo $J$}
        {
            $\rho_j \mathrel{+}= vehicles_{jt}$ \;
        }
        \For{$i \gets 1$ \KwTo $I$}
        {
            $\kappa_i = \sum^{J}_{j=1} \rho_j \cdot  \mathds{1}_{\pmb{\varphi}_i}(\nu_j) $ \;
        }
        \If{$\phi_{green} > \phi_{min}$}
        {
            \If{not $0 < vehicles_{\phi_{green}} < \mu$}
            {
                \If{$\max \kappa > \theta$}
                {
                    action = $\argmax_i \kappa$ \;
                    $\pmb{\rho}^i = \pmb{0}$ \;
                }
            }
        }
    }
\end{algorithm}

Let us introduce $\pmb{\rho}\in\mathbb{N}^{J}$, where $\pmb{\rho} = (\rho_1, ..., \rho_{j})$ is a set of counters which represent the integrals of vehicles per lane over time and $J$ is the number of approaching lanes. Now, when lane $\nu_j$ gets green light, independently which phase turned this light into green, $\rho_j$ corresponding to lane $\nu_j$ is reset to zero.

Then, instead of counting $\kappa_i$ directly, $\kappa_i$ is calculated by summing its corresponding $\rho$-values:
\[\kappa_i = \sum^{J}_{j=1} \rho_j \cdot  \mathds{1}_{\pmb{\varphi}_i}(\nu_j) \]
where $\mathds{1}_{\pmb{\varphi}_i}(\nu_j)$ is a function that indicates whether lane $v_j$ has green light in phase $\pmb{\varphi}_i$.
Note that $\kappa_i$ remains the cumulative waiting time of phase $\pmb{\varphi}_i$, like in the original SOTL method.

Now, whenever a lane $\nu_j$ gets green light, all the corresponding $\kappa$-values are updated accordingly, because they are all dependent on the same $\rho_j$. So, by introducing an additional parameter, passed cars will be removed from all corresponding integrals, and consequently, true to the original SOTL method, the phase with the most waiting time is chosen.

The full algorithm is presented in Algorithm~\ref{sotl}, in which $\phi_{green}$ is the current duration of the phase, $\phi_{min}$ is the minimal duration a phase must be, $vehicles_{\phi_{green}}$ is the number of vehicles within a hand-tuned distance from the green lights, $\mu$ is a tunable parameter indicating the number of vehicles that is needed to split up a cluster and $\pmb{\rho}^i$ is the set of $\rho$ values corresponding to phase $\pmb{\varphi}_i$. Note that in the original two-phase, four-approach setting SOTL and SOTL-2.0 exhibit identical behavior.

%% file: chapters/results.tex
\renewrobustcmd{\bfseries}{\fontseries{b}\selectfont}
\renewrobustcmd{\boldmath}{}
\addtolength{\tabcolsep}{-4.1pt}

        
    
        
        

\renewrobustcmd{\bfseries}{\fontseries{b}\selectfont}
\renewrobustcmd{\boldmath}{}
\newrobustcmd{\B}{\bfseries}
\addtolength{\tabcolsep}{-4.1pt}

In this section, we discuss the experimental setup to evaluate our method. We describe the dataset, model architecture and baseline methods.

\subsection{Data}

We perform experiments on five simulated intersections in Hangzhou, China. The data are based upon real-life traffic recordings and contain two hours of vehicle trajectories per intersection \cite{wei2019colight}. We simulate the traffic in CityFlow\footnote{\url{https://cityflow-project.github.io/}} \cite{zhang2019cityflow} by feeding the route and spawn time of each vehicle and let our RL agent control the traffic lights. Each green light is followed by five seconds of yellow light. Each intersection has eight approaches, four going straight and four turning left. We use all possible phases as actions, i.e. non-conflicting green light configurations. Note that our method can be applied to intersections of any size.

Usually, RL environments are either deterministic or contain some randomness that makes every run different. However, when simulating vehicles based upon a fixed dataset, every run is very similar, while in reality traffic is always different.
Therefore, we split the data into training, validation and test sets to avoid overfitting to a specific hour of traffic flow data, similar to supervised learning. Besides that, using test and validation sets allows us to compare the generalisation of the different methods.

In order to increase generalization to unseen traffic scenarios and simultaneously make maximum use of the available data, we train our model on four intersections and use the two different hours of the remaining fifth intersection for validating and testing. We repeat this for all five intersections.

\subsection{Model Architecture}

To test our framework we use the Deep Q-learning algorithm as proposed by \citet{mnih2015human} with soft-updates \cite{lillicrap2015continuous}. Our neural network consists of two fully-connected hidden layers of 64 nodes with ReLu activation followed by a fully-connected linear layer with a single output for each phase. We use the hyperparameters as proposed by \citet{mnih2015human}, except for using a minibatch size of 512, a learning rate of $1e^{-3}$ and a replay memory size of 360k. We optimize our network with the Adam algorithm 
and use an $\epsilon$-greedy behaviour policy. We use the same network architecture and hyperparameters across all datasets, showing that our approach is robust to a variety of intersections.

\subsection{Baselines}
In addition to the earlier described RL method LIT \cite{zheng2019diagnosing}, we also report scores for a set of rule-based methods to give an intuition about the complexity of the intersection. The method labeled fixed is a cyclic phase order with 20 seconds of green time per phase and the method labeled random is a policy that selects actions uniformly at random. SOTL-1.0 is an implementation\footnote{https://traffic-signal-control.github.io/code.html} of the rule-based Self-Organizing Traffic Lights (SOTL) algorithm \cite{Cools2008}. Note that, although \citeauthor{zheng2019diagnosing} use this implementation throughout their studies, it resembles more the cut-off method designed by \citeauthor{fouladvand2004optimized} than it does resemble the original SOTL method as introduced by \citeauthor{Gershenson2004}. 
SOTL-2.0 is our augmented version of the original SOTL method and picks the phase with the most waiting time, as described in section \ref{chapter:sotl}.

\section{Results}
\label{evaluation}

In this section, we discuss how the performance of our learning algorithm is influenced by choices regarding the state and action representation as well as the nature of the MDP formalisation. We use average travel time as an evaluation metric, which we compute every 50 epochs on the validation set during training. We save the model corresponding to the best score on the validation set and use that model on the test set to compute the final evaluation score. We only perform one run for each experiment, since the low variance between runs, due to the deterministic traffic scenarios used for data generation, does not impact the relative performance of the methods.

\paragraph{State representation} We compare our results with some of the best-performing methods from the ATSC literature, both rule-based and RL-based \cite{Cools2008,zheng2019diagnosing}. The lower four rows of Table~\ref{tab:state_space} show the scores on the validation/test set for each part of our state representation compared to the state representation of LIT. Note how we consistently outperform LIT on all five intersections.

\begin{figure}[tb]
    \centering
    \includegraphics[width=\linewidth]{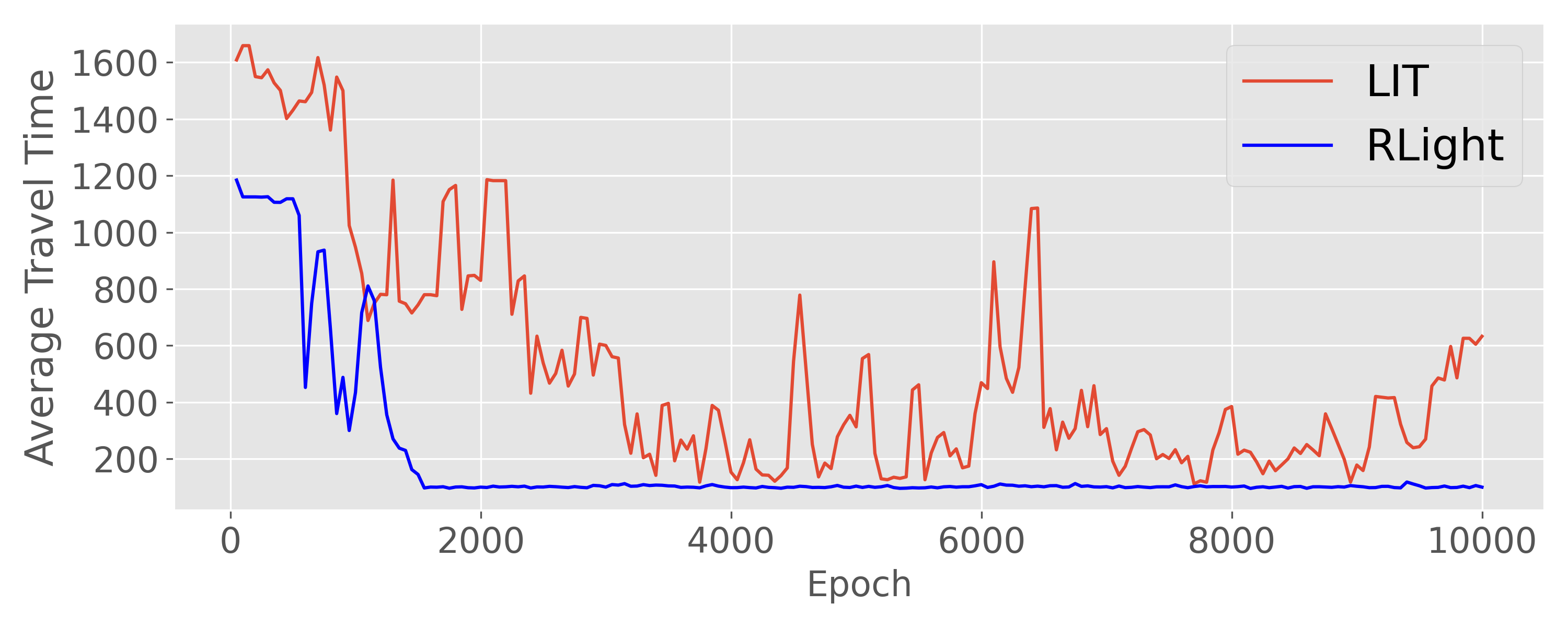}
    \caption{This plot shows how the average travel time per 50 epochs evolves during training on Hangzhou 1 of RLight (blue) compared to LIT (orange). Every 50 epochs correspond to 45000 weight updates or approximately 12 minutes of training time.} 
    \label{fig:traveltime}
\end{figure}

LIT learns policies that are on par with SOTL-2.0 on the validation set but fails to generalize to the test set on intersections 1 and 5. As shown in Figure~\ref{fig:traveltime}, the validation performance of LIT is quite unstable during training, which goes some way towards explaining its poor test time performance. This figure also shows that our streamlining of the framework and consequently of the optimization process has paid out, resulting in very stable learning. 

\begin{table*}[tb]
  \begin{center}
    \caption{The upper table shows the average travel time in seconds for various rule-based methods on the validation/test set. The lower table reports results of each component in our state representation on the validation/test set, while being trained on the other four intersections. $\pmb{w}$ represents the number of waiting vehicles, $\pmb{a}$ the number of approaching vehicles, $\pmb{d}$ the average distance of approaching vehicles and $\pmb{s}$ the average speed of approaching vehicles. Additionally, every state contains the current phase as described in Section~\ref{state_representation}. The experiments are performed with the Semi-Markov Decision Process model.}
    \label{tab:state_space}
    \begin{tabular*}{\textwidth}{@{} l @{\extracolsep{\fill}} *{4}{w{r}{2em} @{\ /\ \extracolsep{0pt}} w{l}{2em} @{\extracolsep{\fill}}} w{r}{2em} @{\ /\ \extracolsep{0pt}} w{l}{2em} @{}}
        \toprule
        & \multicolumn{2}{c}{\makebox[0pt]{Hangzhou 1}} & \multicolumn{2}{c}{\makebox[0pt]{Hangzhou 2}} & \multicolumn{2}{c}{\makebox[0pt]{Hangzhou 3}} & \multicolumn{2}{c}{\makebox[0pt]{Hangzhou 4}} & \multicolumn{2}{c}{\makebox[0pt]{Hangzhou 5}}\\
        \midrule
        Random & 981 & 1086 & 613 & 520 & 649 & 844 & 830 & 1110 & 967 & 1064\\
          Fixed time & 440 & 632 & 283 & 213 & 252 & 278 & 319 & 599 & 539 & 613\\
        SOTL-1.0 & 221 & 335 & 160 & 165 & 117 & 132 & 174 & 290 & 215 & 325\\
          SOTL-2.0 & 120 & 234 & 77 & 78 & 81 & 87 & 96 & 159 & 115 & 193\\
              \midrule
        $[\pmb{w}^\intercal + \pmb{a}^\intercal]$ (LIT) & 112 & 422 & 79 & 73 & 96 & 111 & 100 & 148 & 96 & 842\\
        $[\pmb{w}^\intercal, \pmb{a}^\intercal]$ & 103 & \B 138 & 71 & 74 & 85 & 92 & 88 & 104 & 99 & 116 \\
        $[\pmb{w}^\intercal, \pmb{a}^\intercal, \pmb{d}^\intercal]$ & \textbf{95} & 151 & \B 66 & \B 65 & \B 78 & \B 83 & \B 82 & \B 100 & \B 85 & \B 107 \\
        $[\pmb{w}^\intercal, \pmb{a}^\intercal, \pmb{d}^\intercal, \pmb{s}^\intercal]$ & \B 95 & 142 & 67 & \B 65 & 80 & 87 & \B 82 & \B 100 & \B 85 & 144\\
        \bottomrule
    \end{tabular*}
  \end{center}
\end{table*}

From our different state representations, $[\pmb{w}^\intercal, \pmb{a}^\intercal, \pmb{d}^\intercal]$ achieves the best results on four of the five intersections.
The addition of the average speed $\pmb{s}$ of the cluster of approaching vehicles leads to slightly worse performance, which might indicate that it does not contribute enough to compensate for the expansion in state space.

Additionally, we show that SOTL-2.0 consistently achieves at least 30\% lower travel times than SOTL-1.0, with an average reduction of 38\%. Still, our RLight agent using state representation $[\pmb{w}^\intercal, \pmb{a}^\intercal, \pmb{d}^\intercal, \pmb{p}^\intercal_t]$ outperforms SOTL-2.0 by a substantial margin of at least 34\% on all five intersections despite incorporating almost no prior knowledge about traffic flows.

Note that intersections 1, 4 and 5 are more crowded than the other intersections, which increases the impact of each action and therefore requires more precise control. We show that in these cases SOTL and LIT do not generalize well from the validation set to the test set, while our method excels especially in these cases, achieving a striking average of 75\% lower travel times than LIT.

Finally, we show that the results on the validation and test set vary significantly from each other, suggesting the methods are prone to overfitting to a specific hour of traffic flow at the intersection. This confirms the decision to split the data into train, validation and test sets, which is uncommon in reinforcement learning. This is probably because we generate experiences based upon real data, which only allows for a finite amount of experiences, whereas many reinforcement learning applications allow for the generation of an infinite stream of non-deterministic data.

\paragraph{Action space} We compare the effect of using a cyclic versus acylic phase order in Table~\ref{tab:action_space}. The freedom the agent gets in choosing a phase independently of the last phase improves performance consistently on all intersections. Moreover, selecting a wrong action in cyclic control has a longer effect on the queues since each action determines the next actions, becoming more sensitive to mistakes. Additionally, we show that also when adopting a cyclic phase order, our state representation outperforms LIT's, even when LIT's state representation was specifically aimed towards cyclic phase orders.

\begin{table}[tb]
  \begin{center}
  \caption{The average travel time in seconds of a cyclic and acyclic action space using our state representation $[\pmb{w}^\intercal, \pmb{a}^\intercal, \pmb{d}^\intercal, \pmb{p}^\intercal_t]$ relative to the state representation of LIT. Note that our state representation also outperforms LIT's when adopting a cyclic phase order.}
    \begin{tabular*}{\linewidth}{@{}l@{\extracolsep{\fill}}*{4}{c@{\extracolsep{\fill}}}c@{}}
    \toprule
       & \text{H1} & \text{H2} & \text{H3} & \text{H4} & \text{H5}\\
      \midrule
      Cyclic LIT & 422 & 88 & 119 & 470 & 184\\
      Cyclic RLight & 248 & 78 & 93 & 142 & 142 \\
      \midrule
      Acyclic RLight & \B 151 & \B 65 & \B 83 & \B 100 & \B 107\\
      \bottomrule
    \end{tabular*}
    \label{tab:action_space}
  \end{center}
\end{table}

\paragraph{Markov Decision Process} In Table~\ref{tab:mdp_smdp} we compare modelling the process as a Markov Decision Process (MDP) versus a Semi-Markov Decision Process (SMDP). Both methods result in similar scores, but the SMDP requires 37\% less training time than the MDP, because the replay memory is filled only with valuable timesteps, thereby needing fewer gradient updates relative to the MDP.

\begin{table}[tb]
  \begin{center}
  \caption{The difference in average travel time in seconds of modelling the problem as a Markov Decision Process (MDP) versus a Semi Markov Decision Process (SMDP) using our state representation $[\pmb{w}^\intercal, \pmb{a}^\intercal, \pmb{d}^\intercal, \pmb{p}^\intercal_t]$. The results are similar but the SMDP requires significantly less training time.}
    \begin{tabular*}{\linewidth}{@{}l@{\extracolsep{\fill}}*{4}{c@{\extracolsep{\fill}}}c@{}}
        \toprule
& \text{H1} & \text{H2} & \text{H3} & \text{H4} & \text{H5}\\
      \midrule
        \text{ MDP } & 134 & 68 & 83 & 97 & 101 \\
        \text{ SMDP } & 151 & 65 & 83 & 100 & 107\\

      \bottomrule
    \end{tabular*}
    \label{tab:mdp_smdp}
  \end{center}
\end{table}

\begin{figure*}[ht]
    \centering
    \includegraphics[width=\linewidth]{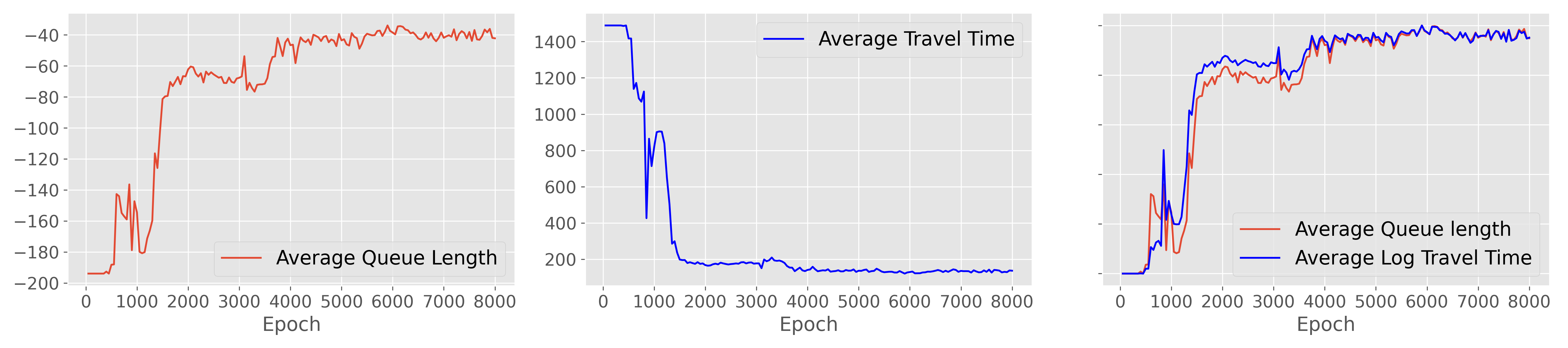}
    \caption{The left and middle plots show the average negative queue length and average travel time respectively during training. The right plot shows the proportionality between these two quantities by depicting the log travel time upside-down.} 
    \label{fig:rewards_traveltime}
\end{figure*}

\paragraph{Reward function} Figure~\ref{fig:rewards_traveltime} visually shows the proportionality of our reward function and the average travel time, which confirms that negative cumulative queue length is a good shaping reward for average travel time.

\paragraph{Generalisation} 


\begin{figure}[ht]
    \centering
    \includegraphics[scale=0.45]{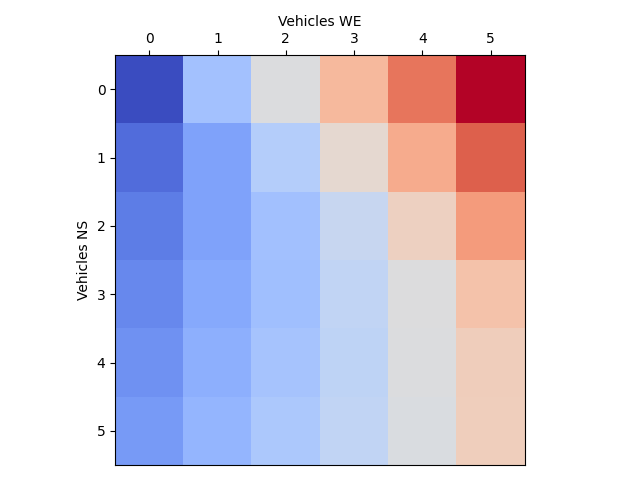}
    \caption{This figure shows the difference between the two Q-values corresponding to switching to WE or keeping phase NS in a simple four-approach two-phase intersection. Warmer colors indicate a switch, cooler colors indicate keeping the phase and grey values indicating indifference. This shows how the decision boundary is fairly symmetrical in the diagonal, indicating an underlying function over the traffic densities. Note how the decision boundary is shifted to the right due to the costs of switching (the addition of yellow light).} 
    \label{fig:generalisation}
\end{figure}

To investigate the generalization we input artificial states with zero to five vehicles per direction into a trained Q network in a simple four-approach two-phase intersection.
We depict the corresponding Q values in Figure \ref{fig:generalisation} to show the decision boundary of choosing one over the other phase. The consistent linearity between the Q values indicates that the model has learned to generalize well to unseen states, which suggests that there exists an underlying function over the traffic densities, independent of a specific intersection. Consequently, we hypothesize that the more diverse the training samples are, the more this underlying function gets explored and learned, resulting in good understanding of where the decision boundary lies. This suggests that training on multiple intersections and augmenting the data could increase generalisation, which consequently could increase performance in the unpredictable traffic scenarios of the real-world.

%% file: chapters/discussion.tex
While our results look promising, we have only tested our approach on data from one source with only two hours of traffic data per intersection. In addition, the layouts of the intersections are identical, although we have designed our method to be applicable to any form of intersection. 



Our results are also dependent on the realism of the simulation. Firstly, all vehicles are simulated with the maximum speed, which unrealistically makes it an undescriptive factor in our RL environment. Secondly, vehicles are unable to collide in the simulator, which should be built in to provide a balanced view on safety.


Our state representation could be extended to incorporate bicycle lanes, which we reckon to be fairly straightforward due to the compact nature of our state representation. Furthermore, in extremely dense urban environments, augmenting the state representation by including the amount of free space on the outgoing lanes could become useful. This is because in these cases not every vehicle might be able to pass through green and accordingly, the agent should change its decisions.

The reward function could potentially be augmented to prioritize certain vehicles (e.g. trucks, public transport, bicycles) by assigning each of them a different weight. Also, each vehicle could be weighted by its waiting time in order to decrease the variance of the travel time at the expense of the average travel time. 

Interesting follow-up research would be to investigate whether our algorithm performs strongly in a multi-agent scenario as well, where several connected intersections are controlled by individual agents. We hypothesize that the natural formation of clusters allows for implicit coordination between intersections, due to our agent's ability to deal with clustered traffic.

%% file: chapters/conclusion.tex
In this work, we have opted to overcome the lack of a straight-forward reinforcement learning framework in traffic signal control by investigating choices regarding the fundamental premises of a deep reinforcement learning agent.
Concretely, we have sought to find ways to exploit relevant inductive biases regarding the state, reward, action and MDP formulation.
We have shown that we can speed up learning by reformulating the MDP as an SMDP by discarding yellow phase time from the learning process, that acyclic phase transitions consistently yield better performance than cyclic phase transitions, and that distinguishing approaching and waiting vehicles is necessary to effectively deal with non-uniform traffic flows. Additionally, we have shown that splitting up the data into train-validation-test sets is appropriate in traffic signal control and that our generalisation of the Self-Organising Traffic Lights leads to a strong baseline in multi-phase intersections.

Our empirical evaluations on real-world datasets have shown that our proposed RLight method produces stable learning and outperforms the chosen baseline methods by a substantial margin. 
These results demonstrate the potential of reinforcement learning methods to improve urban traffic flows, thereby reducing travel time and cutting CO2 emissions.